\documentclass[conference]{IEEEtran}
\IEEEoverridecommandlockouts
\usepackage{cite}
\usepackage{amsmath,amssymb,amsfonts}
\usepackage{algorithm}
\usepackage{algorithmic}
\usepackage{graphicx}
\usepackage{textcomp}
\usepackage{xcolor}
\usepackage{hyperref}
\hypersetup{
    colorlinks=true,
    citecolor=black,
    linkcolor=black,
    filecolor=black,      
    urlcolor=black,
    bookmarks=True,
    pdftitle={Online outlier Detection Threshold},
    pdfauthor={Marek Wadinger}
    pdfpagemode=FullScreen,
    }
\def\BibTeX{{\rm B\kern-.05em{\sc i\kern-.025em b}\kern-.08em
    T\kern-.1667em\lower.7ex\hbox{E}\kern-.125emX}}
\begin{document}

\title{Real-Time Outlier Detection with Dynamic Process Limits\\

}

\author{\IEEEauthorblockN{Marek Wadinger and Michal Kvasnica}
\IEEEauthorblockA{\textit{Institute of Information Engineering, Automation and Mathematics} \\
\textit{Slovak University of Technology in Bratislava}\\
Bratislava, Slovakia \\
\{marek.wadinger, michal.kvasnica\}@stuba.sk}
}

\maketitle

\begin{abstract}
Anomaly detection methods are part of the systems where rare events may endanger an operation's profitability, safety, and environmental aspects. Although many state-of-the-art anomaly detection methods were developed to date, their deployment is limited to the operation conditions present during the model training. Online anomaly detection brings the capability to adapt to data drifts and change points that may not be represented during model development resulting in prolonged service life. This paper proposes an online anomaly detection algorithm for existing real-time infrastructures where low-latency detection is required and novel patterns in data occur unpredictably. The online inverse cumulative distribution-based approach is introduced to eliminate common problems of offline anomaly detectors, meanwhile providing dynamic process limits to normal operation. The benefit of the proposed method is the ease of use, fast computation, and deployability as shown in two case studies of real microgrid operation data.
\end{abstract}


\section{Introduction}\label{Introduction}
The era of Industry 4.0 is ruled by data. Effective data-based decision-making is driven by the quantity of collected data. Internet of Things (IoT) devices made data acquisition seamless and positively influenced a wide range of industries. It is estimated that the annual economic impact of IoT will further grow and reach up to \$6.2 trillion by 2025 \cite{Manyika2013}. 

Various data collection mechanisms are used to buffer and store the data for future processing. However, the tremendous increase in data availability and the desire to extract valuable insight led to problems with the unbounded buffering and storage capacity. Real-time evaluation of the data streams became an acronym for smart data processing. 

Streaming data analytics introduced mechanisms for online extraction and transformation while loading to the storage only a fraction of the former data load, which allowed the storage of the vital information carried by the data more comprehensively. However, the unstable quality of the data appeared to have the most crucial importance over the quantity. 

Anomaly detection, well studied in the last decades, was reborn to the world of new challenges. Former studies were mainly concerned with a domain-specific detection of various anomalies while trained offline \cite{Chandola2009}. However, anomalies of diverse sources, from fraudulent web activity and suspicious financial transactions to sensor failure, malfunctioning of the hardware, and performance drops, mutate over time, and the model had to be updated.

Companies expanded their research activities on the creation and integration of generic frameworks combining prediction, detection, and alert mechanisms. One of the first projects, open-sourced for the public, are EGADS by Yahoo \cite{Laptev2015} and AnomalyDetection by Twitter \cite{Kejariwal2015}. The frameworks' modularity allowed the automation of the anomaly detection of time-series data and created space for discussion. 

Moving from domain-specific to generic methods posed new problems connected to type I errors, i.e., a false-positive classification of normal behavior as anomalous. Accurate selection of forecaster, detector, and alerting mechanism allowed to tackle the problem, nevertheless, introduced considerable dependence on expert domain knowledge and fine-tuning. 

Further work proved improvement in performance while relieving the tight requirements on domain knowledge \cite{Ahmad2016}. However, strict demands on detection systems ranging from lasting up times to continuous monitoring with stable performance pointed to the challenge of data stationarity. Change points and concept drifts troubled unsupervised models, which led to service downtime due to the model retraining.  

The era of adaptive machine learning introduced incremental learning schemes as a solution. Multiple studies for learning modes, adaptation methods, and model management swept through the machine learning community. Pannu et al. proposed an adaptive anomaly detection system \cite{Pannu2012}. However, the method represented a supervised operator-in-the-loop solution. Zhang et al. introduced an adaptive kernel density-based algorithm that uses an adaptive kernel width \cite{Zhang201850}. Nonetheless, training the models on big data had limitations resulting from the storage and unbounded buffering of data. Online learning models relaxed the need for data availability during model training \cite{Gama2014}. On the contrary, it processed the data from a bounded buffer sequentially as in \cite{Bosman201514} and \cite{Ahmad2017134}. 

Anomaly detection in microgrids, however, called for low latency detection which implied real-time training and prediction processes \cite{Liu2017}. Such adaptation of streamed modeling took into consideration strict boundaries on computational time. For work in this area see \cite{Wang2020114145} and \cite{Dai2022}.

Alerting mechanisms in process automation detect situations where signal value deviates from constraints. An alert watchdog is triggered on threshold violation by individual signals. The constraints, or process limits, are usually predefined and fixed. Nevertheless, factors such as aging and environmental changes call for dynamic process limits. Setting up a procedure for an evergrowing number of signal measurements is time-consuming. Besides, it is impossible for signals where no prior information about a correct process range is known. Those are subject to external factors that are unknown at setup time. 

In this article, we suggest using existing process automation infrastructure based on alerting (PLC, SCADA, among others) and applying machine learning for dynamic process range based on changing conditions. We propose an unsupervised anomaly detection algorithm capable of online adaptation to change points and concept drifts, which adds to a recently developed body of research. The approach is evaluated on two case studies of microgrid sensors. To the author's knowledge, there are no studies to date concerned with providing adaptive operation constraints.

The main benefits of the proposed solution are that it:
\begin{itemize}
\item Keeps existing IT infrastructure, saving costs, and does not require operator retraining
\item Automates alerting thresholds setup for a high number of signals
\item Automates alerting for signals with no a priori knowledge of process limits
\item Assesses changing environmental conditions and device aging
\item Uses self-learning approach on streamed data
\end{itemize}


\section{Preliminaries}
This section introduces the main concepts which are building pillars of the developed approach. Subsection \ref{AA:Welford} will discuss a one-pass algorithm that allows for online adaptation. The following Subsection \ref{AA:InvWelford} proposes the ability to invert the solution in a two-pass implementation. The mathematical background of distribution modeling in Subsection \ref{AA:Distribution} provides a basis for the Gaussian anomaly detection model conceptualized in the last Subsection \ref{AA:Anomaly} of Preliminaries.

\subsection{Welford's Method}\label{AA:Welford}
Streaming data analytics, restrict the unbounded buffer or storage of the data, i.e., limits the uncontrolled growth of memory usage with the increasing amount o input data. In such cases, it is desired to keep the data only for the period of time required to perform computations. For the given purpose serve one-pass algorithms. This category of methods allows processing on-the-fly without the need to store the entire data stream. 
\newtheorem{definition}{Definition}[section]
\begin{definition}[One-pass algorithm]
The algorithm with a single access to the data items in the order of their occurrence, i.e., \(x_1,x_2,x_3,...\) is called one-pass algorithm \cite{Schw09}
\end{definition}

Welford's method represents a numerically stable one-pass solution for the online computation of mean and variance \cite{Wel62}. 
Given \(x_i\) where \(i=1,...,n\) is the sample index in given population  \(n\), the corrected sum of squares is defined as
\begin{equation}
S_n = \sum_{i=1}^n (x_i - \bar x_n)^2\text{,}\label{eq:sumsquares}
\end{equation}
where the running mean \(\bar x_n\) is
\begin{equation}
\bar x_n = \frac{n-1}{n} \bar x_{n-1} + \frac{1}{n}x_n = \bar x_{n-1} + \frac{x_n - \bar x_{n-1}}{n}\text{.}\label{eq:runmean}
\end{equation}
The following identities to update the corrected sum of squares hold true
\begin{equation}
S_n = S_{n-1} + (x_n - \bar x_{n-1})(x_n - \bar x_n)\text{,}\label{eq:upsumsquares}
\end{equation}
and the corresponding variance is
\begin{equation}
s^2_n = \frac{S_{n}}{n-1}\text{.}\label{eq:runvar}
\end{equation}

As we can see in \eqref{eq:upsumsquares}, we do access only current data sample \(x_n\) and previous value of \(\bar x_{n-1}\) which is updated in \eqref{eq:runmean} using the same data sample and the size of seen population \(n\).

\subsection{Inverse Welford's Method}\label{AA:InvWelford}
Let the incoming stream of data be subject to the concept drift. Such alternation in statistical properties has a negative influence on prediction accuracy. Therefore, an adaptation of any machine learning model is crucial for successful long-term operation. 

\begin{definition}[Concept drift]
Concept drift is a change in the statistical properties that occur in a sub-region of the feature space.
\end{definition}

The previous Subsection \ref{AA:Welford} defined the main concept of online statistical computation that allows reacting to such changes. However, the further in time the shift occurs, the slower the adjustment of the running mean is, resulting from a negative relationship in \eqref{eq:runmean} between population size \(n\) and influence of the last sample in population \(x_n\) on the updated value of \(\bar x_{n}\). For this reason, we define the expiration period \(t_e\), over which the running statistics are computed. After the expiration period, the data items are forgotten. Such reversal results in a need to store all the data in the window in order to revert their effect. Given \(t_e=n-1\) we can revert the influence of the first data sample on the running mean as

\begin{equation}
\bar x_{n-1} = \frac{n}{n-1} \bar x_{n} - \frac{1}{n-1}x_{n-t_e} = \bar x_{n} - \frac{x_{n-t_e} - \bar x_{n}}{n-1}\text{,}\label{eq:revmean}
\end{equation}

then reverting the sum of squares follows as

\begin{equation}
S_{n-1} = S_n - (x_{n-t_e} - \bar x_{n-1})(x_{n-t_e} - \bar x_n)\text{,}\label{eq:revrunmean}
\end{equation}

which allows the computation of variance 

\begin{equation}
s^2_{n-1} = \frac{S_{n-1}}{n-2}\text{.}\label{eq:revvar}
\end{equation}

\subsection{Modeling Distribution}\label{AA:Distribution}
Statistical distribution can be used to create a generalized model of a normal system behavior based on observed measurement. Specifically, if no change point is expected in a given subset of samples, the Gaussian normal distribution can be fitted. Parameters of the normal distribution are used to compute  standard score \eqref{eq:zscore} for each new observation. 

\begin{definition}[Standard Score]
Standard score or \(Z\)-score is a number that specifies the number of sample standard deviations \(s^2_n\) by which observation \(x\) deviates from the sample mean \(\bar x_n\) of normal distribution
\begin{equation}
z_n = \frac{x_n - \bar x_n}{s^2_n}\text{.}\label{eq:zscore}
\end{equation}
\end{definition}

In order to define the general probability of \(z\)-score belonging to anomaly we use probability computed using Cumulative Distribution Function (CDF). However, the \(z\)-score must be bounded using an error function into the interval from 0 to 1.

\begin{definition}[Approximate Error Function]
The approximate error function represents the approximate probability that the random variable \(X\) lies in the range of \([\,-z_n,z_n]\,\) denoted as
\begin{equation}
E_A (z_n) = z_n\frac{e^{-z^2_n}}{\sqrt{\pi}}( \,2/1 + 4/3x^2 + 8/15 x^4 + ...) \,\text{.}\label{eq:erf}
\end{equation}
\end{definition}

\begin{definition}[Cumulative Distribution Function (CDF)]
CDF represents the probability that the random variable \(X\) takes a value less than or equal to \(x_i\). \(F_X\colon \mathbb{R} \to [0,1]\). For generic normal distribution with sample mean \(\bar x_n\) and sample deviation \(s_n\) the cumulative distribution function \(F_X(x)\) equals to
\begin{equation}
F_X(x_i)_n = \frac{1}{2}( \,1+E_A(\,\frac{z_n}{\sqrt{2}})\,) \text{.}\label{eq:cdf}
\end{equation}
\end{definition}

Given the probability, we can also derive the value of \(x\) to which it belongs using a percent point function to compute inverse CDF (ICDF) denoted also as $F_X(x_i)^{-1}_n$.

\begin{definition}[Percent-Point Function (PPF)]
PPF returns the threshold value for random variable \(X\) under which it takes a value less than or equal to the value, for which \(F_X(x)\) takes probability lower than selected quantile \(q\). \(Q_X\colon [0, 1] \to \mathbb{R}\). An algorithm that calculates the value of the PPF is reported below as Algorithm \ref{alg:ppf}.
\end{definition}

\begin{algorithm}[H]
\caption{{Percent-Point Function for Normal Distribution}} \label{alg:ppf}
 \begin{algorithmic}[1]
 \renewcommand{\algorithmicrequire}{\textbf{Input:}}
 \renewcommand{\algorithmicensure}{\textbf{Output:}}
 \REQUIRE quantile $q$, sample mean $\bar x_n$ \eqref{eq:runmean}, sample variance $s^2_n$ \eqref{eq:runvar}
 \ENSURE  threshold value $x_{n,q}$
 \\ \textit{Initialisation} : 
  \STATE $f \leftarrow 10$; $l \leftarrow -f $; $r \leftarrow f;$
 \\ \textit{LOOP Process}
  \WHILE {$F_X(l)-q > 0$}
  \STATE $r \leftarrow l $;
  \STATE $l \leftarrow lf $;
  \ENDWHILE
  \WHILE {$F_X(r)-q < 0$}
    \STATE $l \leftarrow r $;
    \STATE $r \leftarrow rf $;
  \ENDWHILE
  \STATE {$\tilde{x}_{n,q} = \text{arg} \min_z \| F_X(z) - q \| ~ \text{s.t.} ~ l \le z \le r$}
 \RETURN $\tilde{x}_{n,q}  \sqrt{s^2_n} + \bar x_n $
 \end{algorithmic}
\end{algorithm}

\subsection{Gaussian Anomaly Detection}\label{AA:Anomaly}
Anomalies come in various kinds and flavors. Commonly denoted types are point (spatial), contextual, and collective (temporal) anomalies \cite{Chandola2009}.
Spatial anomalies take on a value that particularly deviates from the sample mean \(\bar x_n\). From a statistical viewpoint, spatial anomalies can be considered values \(x\) that significantly differ from the data distribution. 

In empirical fields, such as machine learning, the three-sigma rule defines a region of distribution where normal values are expected to occur with near certainty. This assumption makes approximately 0.27\% of values in the given distribution considered anomalous. 

\begin{definition}[Three-Sigma Rule of Thumb (3\(\sigma\) rule)]
3\(\sigma\) rule represents a probability, that any value \(x_i\) of random variable \(X\) will lie within a region of values of normal distribution at the distance from the sample mean \(\mu_n\) of at most 3 sample standard deviations \(\sigma_n\).
\begin{equation}
P\{|x_i-\mu_n|<3\sigma_n\}=0.99730
\end{equation}
\end{definition}

Anomalous values occur on both tails of the distribution. In order to discriminate the anomalies using the three-sigma rule on both tails of the distribution, we define the anomaly score as follows

\begin{equation}
y_i = 2 \left|{F_X(x_i)_n - \frac{1}{2}}\right|\text{,}\label{eq:score}
\end{equation}

where 
\begin{subequations}

\begin{align}
y_i \in [0,P\{|x_i-\mu_n|<3\sigma_n\})\text{,}\label{eq:score_norm}
\end{align}

applies for normal observations and 

\begin{align}
y_i \in [P\{|x_i-\mu_n|<3\sigma_n\},1]\text{,}\label{eq:score_anomaly}
\end{align}
\end{subequations}
 for anomalies.

Using pure statistics to model normal behavior lets us ask the question about the threshold value \(x\) which corresponds to the area under the curve of CDF equal to the given probability. A such query can be answered using inversion of \eqref{eq:score}. However, inversion of \eqref{eq:score} would fail the horizontal line test. Therefore, we restrict the applicability of the inverse only to \(F_X(x)_i \in [0.5, 1]\)

\begin{equation}
x_i = F_X\left(\frac{y_i}{2}+\frac{1}{2}\right)^{-1}_n\label{realthresh}
\end{equation}

In order to derive a lower threshold, the Gaussian distribution is fitted to the negative value of the streamed data and evaluated accordingly using the previously defined equations.

\section{ICDF-based Real-Valued Threshold System}
We suggest a novel approach to provide dynamic process limits using an online outlier detection algorithm capable of handling concept drifts in real-time. Our main contribution is based on using an inverse cumulative distribution function (ICDF) to supply a real-valued threshold for anomaly detection, i.e., to find the values of the signal which corresponds to the alert-triggering process limits. Therefore, in the context of machine learning, we are tackling an inverse problem, i.e., calculating the input that produced the observation. To utilize an adaptive ICDF-based threshold system, the univariate Gaussian distribution has to be fitted to the data in online training and ICDF evaluated on the fly. This method is divided into four parts and described in the following lines. For a simplified representation of the method see Algorithm \ref{alg:detector}.

\subsection{Model Initialization}\label{init}
The initial conditions of the model parameters are \(\mu_0 = x_0\) for mean and \(s^2_0 = 1\) for variance. The score threshold is constant and set to \(q = 0.9973\). Moreover, there are two user-defined parameters: the expiration period $t_e$, and the time constant of the system $t_c$. The expiration period, which defines the period over which the time-rolling computations are performed, can be altered to change the proportion of expected anomalies and allows relaxation (longer expiration period) or tightening (shorter expiration period) of the thresholds. The time constant of the system determines the speed of change point adaptation as it influences the selection of anomalous points that will be used to update the model for a window of values \(Y=\{y_{i-t_c},...,y_{i}\}\) if the following condition holds true
\begin{equation}
{\sum_{y\in Y}y \over n(Y)} > q\text{.}\label{eq:condition}
\end{equation}
The existence of two tunable and easy-to-interpret hyper-parameters makes it very easy to adapt the solution to any univariate anomaly detection problem.

\subsection{Online training}\label{train}
Training of the model takes place in an online fashion, i.e., the model learns one sample at a time at the moment of its arrival. Learning updates the mean and variance of the underlying Gaussian distribution. The computation of moving mean \eqref{eq:runmean} and variance \eqref{eq:runvar} is handled by Welford's method. Each sample after the expiration period is forgotten and its effect reverted in the second pass. First, the new mean is computed using \eqref{eq:revmean} which accesses the first value in the bounded buffer. The value is dropped in the same pass. Second, the new sample variance is reverted based on \eqref{eq:revvar} using the new mean and current mean that is overwritten afterward. For details see Subsection \ref{AA:InvWelford}.

\subsection{Online prediction}\label{predict}
In the prediction phase, \(z\)-score \eqref{eq:zscore} is computed and passed through $E_A$ \eqref{eq:erf} in order to evaluate $F_{X}(x_i)$ from \eqref{eq:cdf}. The algorithm marks the incoming data points if their corresponding anomaly score \eqref{eq:score} is out of the range defined by threshold \(q\). In other words, marks signal value \(x_i\) that is higher or equal to the threshold, which bounds the three-sigma region.

\subsection{Dynamic Process Limits}\label{constrait}
Normal process operation is constrained online using ICDF. The constant value of \(q\) and parameters of the fitted distribution are both passed through Algorithm \ref{alg:ppf} to obtain value, which corresponds to the value of \(x\) that would trigger an upper bound outlier alarm at the given time instance. To obtain a lower bound of operation conditions the same procedure is applied to the distribution fitted on negative values of input.

\begin{algorithm}[H]
\caption{{Online Anomaly Detection Workflow}} \label{alg:detector}
 \begin{algorithmic}[1]
  \renewcommand{\algorithmicrequire}{\textbf{Input:}}
  \renewcommand{\algorithmicensure}{\textbf{Output:}}
  \REQUIRE expiration period $t_e$, time constant $t_c$
  \ENSURE  score $y_i$, threshold $x_{i,q}$
 \\ \textit{Initialisation} : 
  \STATE $i \leftarrow 1;~ n \leftarrow 1;~ q \leftarrow 0.9973;~ \bar x  \leftarrow x_0;~  s^2 \leftarrow 1$;
  \STATE compute $F_X(x_0)$ using \eqref{eq:zscore};
 \\ \textit{LOOP Process}
  \LOOP
    \STATE {$x_i \leftarrow$ RECEIVE()};
    \STATE $y_i \leftarrow$ PREDICT($x_i$) using \eqref{eq:score};
    \STATE $x_{i,q} \leftarrow$ GET($q, \bar x, s^2$) using Algorithm \ref{alg:ppf};
    \IF {\eqref{eq:score_norm} \OR \eqref{eq:condition}}
     \STATE {$\bar x$, $s^2 \leftarrow$ UPDATE($x_i, \bar x, s^2, n$) using \eqref{eq:runmean}, \eqref{eq:runvar}};
     \STATE $n \leftarrow n + 1$;
     \FOR {$x_{i-t_e}$}
      \STATE {$\bar x$, $s^2 \leftarrow$ REVERT($x_{i-t_e}, \bar x, s^2, n$) using \eqref{eq:revmean}, \eqref{eq:revvar}};
      \STATE $n \leftarrow n - 1$;
     \ENDFOR
    \ENDIF
    \STATE $i \leftarrow i + 1$;
  \ENDLOOP
 \end{algorithmic}
\end{algorithm}

\section{Case Study}
In this section, we demonstrate the applicability of the proposed ICDF-based approach in two case studies of the microgrid operation. The properties and performance were investigated using streamed signals from the IoT devices. The successful deployment demonstrates that this approach is suitable for existing alerting mechanisms of process automation infrastructure.

The case studies were realized using Python 3.10.1 on a MAC with an M1 CPU and 8 GB RAM. The percent point function was solved using an iterative root-finding algorithm, Brent's method.

\subsection{Battery Energy Storage System (BESS)}\label{AA}
First, we verify our proposed method on BESS. The BESS reports measurements of State of Charge (SoC), supply/draw energy events, inner temperature, outer temperature, Heating, ventilation, and air conditioning (HVAC) state. Tight control of the battery cell temperature is needed for the optimal performance and maximum lifespan of the battery. Identifying anomalous events and removal of corrupted data might yield significant improvement on the process control level. 

The sampling rate of the signal measurement is 1 minute. However, network communication is prone to packet dropout, which results in non-uniform sampling. To protect the sensitive business value of the data, we normalize all signals to the range $[0, 1$]. The goal was to mark anomalous events in the data and provide adaptive process limits from the online self-learning model. 

\begin{figure}[htbp]
\centerline{\includegraphics{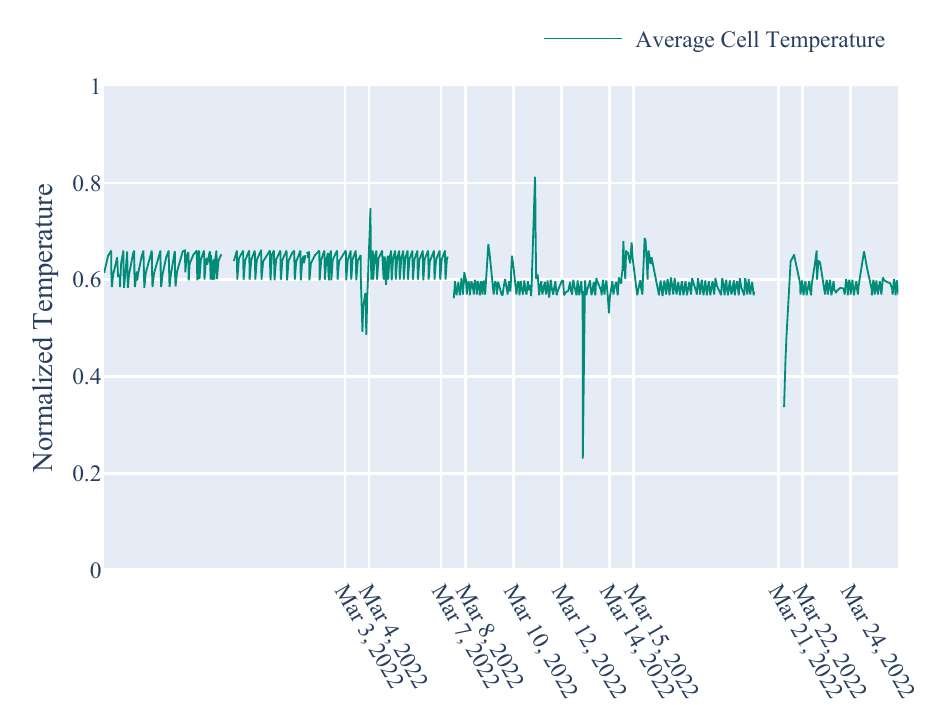}}
\caption{Time Series of Average Battery Cell Temperature measurement (green line). Non-uniform ticks on the x-axis mark days of interest (NOTE: some marks are hidden due to the readability). The y-axis renders the normalized temperature.}
\label{fig:signal}
\end{figure}

Fig. \ref{fig:signal} renders measurement of average battery cell temperature from 21\textsuperscript{st} February until 26\textsuperscript{th} March. We can observe multiple anomalies of various sources given this span, for instance, packet dropout, suspicious events, intermittent sensor failure, and change point in data distribution. Dates of observation given the listed events will be provided later in the paper. 

The initial conditions of the model states are set based on Subsection \ref{init}. The user-defined parameters, were set to 7 days for the expiration period and 5 hours for the time constant. Anomalies found during the first day of the service are ignored due to the initialization of the detector. In this case study, the anomaly detection problem was approached by the online model fitting based on Subsection \ref{train}

\begin{figure}[htbp]
\centerline{\includegraphics{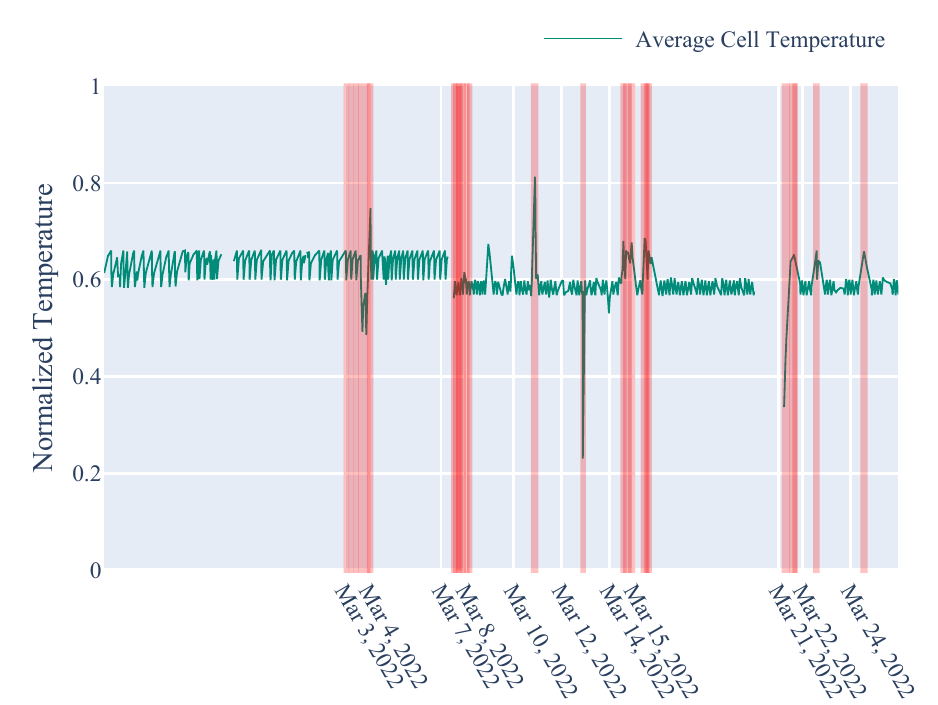}}
\caption{Time Series of Average Battery Cell Temperature measurement (green line) and predicted anomalous events (red vertical rectangles).}
\label{fig:anomalies}
\end{figure}

Using the online prediction described in Subsection \ref{predict} we tag the sample as the anomaly or normal data point. Fig. \ref{fig:anomalies} renders vertical rectangles over the regions from the start until the end of the predicted anomalous event. 

The results on Average Cell Temperature in Fig. \ref{fig:anomalies} show that the model could capture anomalous patterns of various sources. Despite self-learning without supervision, the model-classified anomalies were also confirmed by the data provider after inspection. For instance, a rare event of manipulation with BESS on 3\textsuperscript{rd}, followed by peak on 4\textsuperscript{th} March. BESS relocation on 7\textsuperscript{th}, led to a change point which was alerted and the system adapted completely over the course of 1 day. Test events resulted in peak values through 10\textsuperscript{th} to 15\textsuperscript{th} March, and faulty measurements on the 12\textsuperscript{th} March followed by a packet loss on 21\textsuperscript{st} March were alerted too. The system tagged the next two tests of temperature control switch-offs. 

\begin{figure}[htbp]
\centerline{\includegraphics{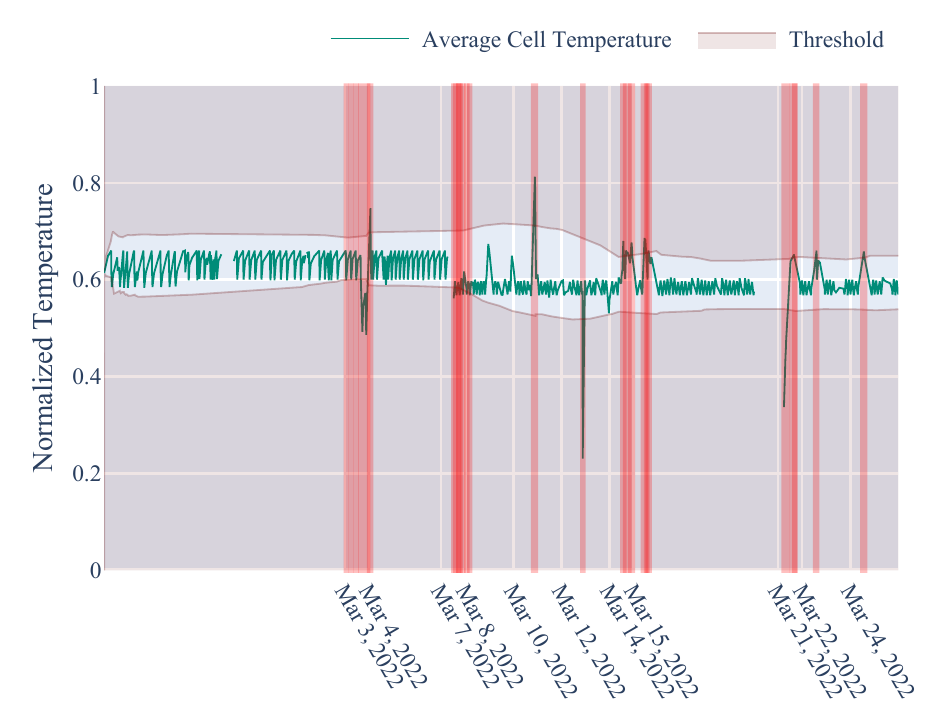}}
\caption{Time Series of Average Battery Cell Temperature measurement (green line) and predicted anomalous events (red vertical rectangles). The reddish fill bonded by the red line represents an area of anomalous behavior as given by the anomaly detector.}
\label{fig:threshold}
\end{figure}

Findings that favor the model's ability to discriminate anomalous behavior are important for the meaningful realization of the dynamic process thresholding. The real-valued threshold mechanism, defined in Subsection \ref{constrait}, provided up-to-date upper and lower bounds for the signal. As for the validity of the dynamic process limits, each breakout of the signal value from within the range was also marked by the anomaly detection system. Fig. \ref{fig:threshold} points to the capability to adapt to change point on 7\textsuperscript{th} March and mitigate the influence of intermittent effects of anomalies on distribution. The speed of the change point adaptation as well as the mitigation of the effect of anomalies on the tightness of limits are governed by the user-defined expiration period and time constant of the system.

\subsection{Power Inverter}\label{AA}
A second case study demonstrates the  proposed method's applicability to the temperature of the power inverter. During high load periods, inverters can heat up swiftly. Technical documentation of every inverter provides details on continuous output rating as a function of temperature that implies static process limits. Normally, for high temperatures, the rating drops rapidly. Nevertheless, the impact of aging and ambient conditions may render conservative limits impractical. Thus the alerting mechanism for the detection of abnormal heating shall be developed. Providing a real-valued anomaly threshold tightens the theoretical operating conditions and gives the ability to track the performance and deviations. 

\begin{figure}[htbp]
\centerline{\includegraphics{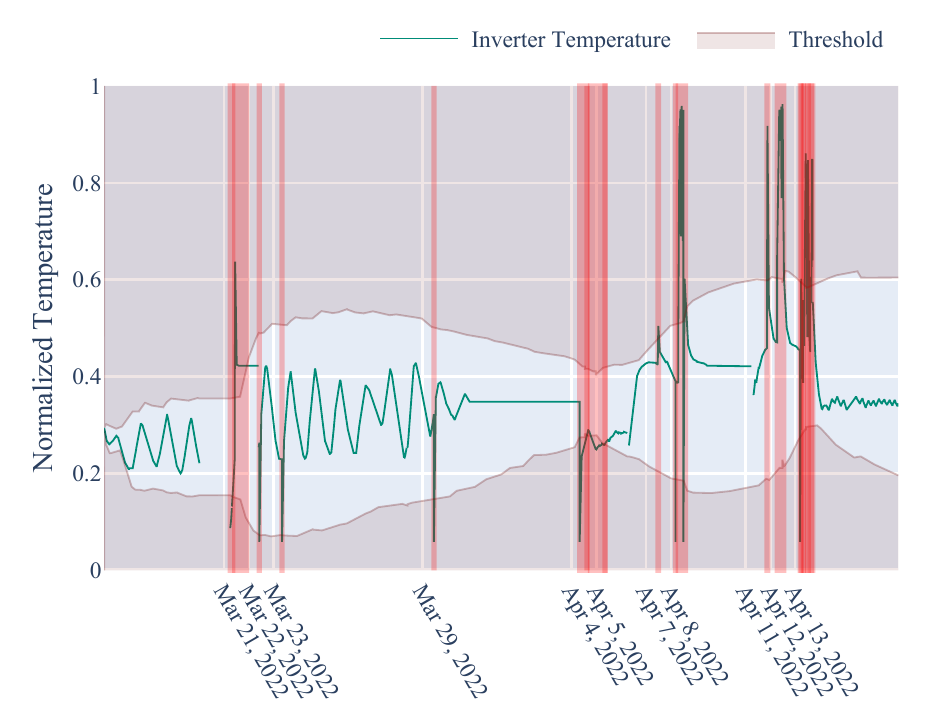}}
\caption{Time Series of Inverter Temperature measurement (green line) and predicted anomalous events (red vertical rectangles). The reddish fill bonded by the red line represents an area of anomalous behavior.}
\label{fig:cs2_threshold}
\end{figure}

Fig. \ref{fig:cs2_threshold} depicts one month of operation of the inverter from 16\textsuperscript{th} March to 17\textsuperscript{th} April 2022. After the packet loss before 21\textsuperscript{st} March, rare temperature events occurred. Both events fell out of the normal operating conditions given by the dynamic process limit. Four faulty sensor readings follow from 22\textsuperscript{nd}, 23\textsuperscript{rd}, 29\textsuperscript{th} March and 4\textsuperscript{th} April. The first two are tagged as anomalies, though almost missed due to the prolonged data loss. Given a shorter time from initialization than $t_e$ the influence of the edge between drop and raise had a relaxing effect on limits. Former finding proposes a need for grace period modification, which would alter self-learning until the buffer given by $t_e$ is not fully filled. The third faulty reading was tagged without influencing the distribution and operational boundaries due to the effect of $t_c$. Oscillations, that kept the boundaries relaxed vanished after 29\textsuperscript{th} March, which further tightened the process limit range. After the fourth caught fault which was not used to update the model, the detector deliberately adapted the range of normal operation during the next day. Outliers during the sensors rescaling period from 7th April were all tagged. However, the relaxed operational conditions would probably lead to ignorance of smaller anomalous oscillations in given period.

\section{Conclusion}
This paper proposes a novel approach to real-time anomaly detection that provides a physical threshold that bounds normal process operation. Such an approach has wide applicability in all the process automation fields where low latency evaluation and online adaptation are crucial. Moreover, adaptive operation constraints provide less conservative process limits and govern important insight into systems behavior. The plug-and-play feature of the model makes it easily deployable as shown in two case studies. 

The first case study performed on BESS examined the average battery cell temperature and demonstrated the ability to capture anomalies as well as the capacity to restrict the operational area by inversion of the cumulative distribution function. Following our investigation of state-of-the-art online anomaly detection described in Section \ref{Introduction} we conclude, that although the robustness and performance of complex methods may exceed the performance of the proposed method, the ability to invert the prediction to depict real-time operational restrictions and eschew using non-comprehensible parameters makes it superior for a wide range of use cases. However, the performance might be greatly afflicted when the time constraints of the observed system are not known. This restriction is much weaker than the restriction of the need for data scientists skilled in the hyper-parameter tuning of unsupervised models. Moreover, hyper-parameter tuning calls for ground truth information about anomalies, which requires an exhaustive collection and is not possible in real time. 

Future works on the method will follow three practical challenges:
Firstly, the multivariate online anomaly detection based on the developed method will be researched. The multivariate implementation would allow the detection of temporal anomalies and the use of features that render spatio-temporal characteristics of the modeled system. This is the common property of most of the online anomaly detection methods that do not offer real-valued thresholds on operational conditions. The multivariate clusters can reveal regions of normal operation that would be otherwise detected incorrectly.

Secondly, the challenge of varying positive and negative process limits thresholds will be examined. As depicted in Fig \ref{fig:cs2_threshold} the positive and negative outliers, in many cases, result from different mechanisms that caused them. The current approach draws a range of normal operational conditions centered around the moving mean value.

Thirdly, automated system identification using normal operation data would further simplify the usage by removing the requirement for system dynamics knowledge. The usage of normal distribution makes the three-sigma rule constrain the number of anomalies only theoretically. This allows the number of anomalies in a given time window to vary greatly and thus the performance is not very sensitive to the selection of the threshold. On the contrary, the time window impacts the model's performance.

\bibliographystyle{IEEEtran}
\bibliography{main}

\end{document}